# Cycle-Consistent Adversarial GAN: the integration of adversarial attack and defense

**Lingyun Jiang[1], Kai Qiao[1], Ruoxi Qin[1], Linyuan Wang[1], Jian Chen[1], Haibing Bu[1], Bin Yan[1]***

[1] National Digital Switching System Engineering and Technological Research Center, Zhengzhou, China

**\* Correspondence:**
Dr. Bin Yan
tttocean_tl@hotmail.com

## Abstract

In image classification of deep learning, adversarial examples where inputs intended to add small magnitude perturbations may mislead deep neural networks (DNNs) to incorrect results, which means DNNs are vulnerable to them. Different attack and defense strategies have been proposed to better research the mechanism of deep learning. However, those research in these networks are only for one aspect, either an attack or a defense, not considering that attacks and defenses should be interdependent and mutually reinforcing, just like the relationship between spears and shields. In this paper, we propose Cycle-Consistent Adversarial GAN (CycleAdvGAN) to generate adversarial examples, which can learn and approximate the distribution of original instances and adversarial examples. For CycleAdvGAN, once the Generator $A$ and $D$ are trained, $G_A$ can generate adversarial perturbations efficiently for any instance, so as to make DNNs predict wrong, and $G_D$ recovery adversarial examples to clean instances, so as to make DNNs predict correct. We apply CycleAdvGAN under semi-white box and black-box settings on two public datasets MNIST and CIFAR10. Using the extensive experiments, we show that our method has achieved the state-of-the-art adversarial attack method and also efficiently improve the defense ability, which make the integration of adversarial attack and defense come true. In additional, it has improved attack effect only trained on the adversarial dataset generated by any kind of adversarial attack.

## 1    INTRODUCTION

With the Deep Neural Networks (DNNs) rapid development, they have achieved great successes in various tasks handling the image recognition[1], text processing[2] and speech recognition[3]. Despite the great success, DNNs have been proved to be vulnerable and susceptible to adversarial example[4], the carefully crafted samples looking similar to natural images but designed to mislead a pretrained model. On the one hand, adversarial example leads to potential security threats by attacking or misleading the practical deep learning applications, for example mistaking a stop sign for a yield sign[5] when auto driving, and a thief for a staff when face recognition[6]. On the other hand, adversarial examples are also valuable and beneficial to not only the deep learning models but also the machine learning model, as they can enhance the robust of models and provide insights into their strengths, weaknesses, and blind-spots[7].

# CycleAdvGAN: integration of adversarial attack and defence

The strategy to generate adversarial examples is to intentionally add imperceptible perturbations to clean instances, for fooling DNNs to make wrong predictions. In the past years, various attack algorithms have been developed to produce adversarial examples in the white-box manner with the knowledge of the structure and parameters of a given model, including gradient based algorithms such as fast gradient sign method[8] and iterative variants of gradient-based methods[9], optimization-based methods such as box-constrained LBFGS[4] and Carlini and Wagner Attacks[10], and network-based methods such as AdvGAN[11] and Natural GAN[12]. At the same time, defense algorithms have progress with advances in attack algorithms, including adversarial training that retrains a neural network to predict correct labels for adversarial examples[8], defensive distillation that migrate knowledge of complex networks to simple networks[13], and network-based defense that add additional network to detecting or denoising[14]. Network-based techniques have achieved satisfying performance not only in terms of attacks but also defenses owning to their great power for generating high-quality synthetic data and detecting these subtle differences to distinguish between adversarial and clean examples. However, existing attack methods exhibit low efficacy when attacking black-box models. For example, most of existing methods, such as FGSM and optimization methods, cannot successfully attack them in the black-box manner, due to the poor transferability[15].Meanwhile, existing defense methods also exist poor transferability, as they only can defense certain attack method. On the other hand, in the previous white-box attacks, the adversary needs to have white-box access to the architecture and parameters of the model all the time.

In addition, those study in these networks are only for one aspect, either an attack or a defense, not considering that attacks and defenses should be interdependent and mutually reinforcing, just like the relationship between spears and shields. Inspired by the idea of GAN, just as the generative model is pitted against an adversary, if then we train a model consisted of attack and defense, and the attacker and defender are also fighting against each other all the time, it can improve the attack ability as well as the defense ability. Besides, it can also improve the transferability, because of better learning the latent distribution of adversarial examples and clean instances. Consequently, utilizing the cycle-consistence idea of CycleGAN[16], we apply a similar paradigm to combine the attacker and defender, which promote each other. In this paper, we propose to train a Cycle-Consistent Adversarial GAN(CycleAdvGAN) to achieve the integration of adversarial attack and defense. Once trained, there is no need to access to the target model itself anymore no matter in the manner of attack or defense, and the CycleAdvGAN can generate perturbations and recovery the adversarial examples such that the resulting example must be realistic according to a discriminator network.

To evaluate the effectiveness of our strategy CycleAdvGAN, we experiment on different datasets including MNIST and CIFAR10 for an ensemble of target models. We evaluate these attack strategies in both semi-white box and black-box settings. We show that adversarial examples generated by CycleAdvGAN have higher success rates in both semi-white box and black-box attacks and also has good effect in defense, due to the fact that attack and defense promote each other, better leaning the distinguish between adversarial example and clean example. In summary, we make the following contributions as follows:

- We are the first to achieve the integration of adversarial attack and defense by preserving a high success rate of attacks as well as defenses.
- We demonstrate a powerful capability of transferability no matter in the manner of attack or defense.
- We indirectly demonstrate the adversarial and clean data are not twins, subjecting to two different distributions.

## 2 RELATED WORK

### 2.1 Adversarial attack with GAN

A number of methods been proposed can successfully generate adversarial examples in the white-box manner, where the adversary has full access to the classifier. A more straightforward approach is to change pixels value simultaneously in the direction of the gradient. And another method is to utilized Generative Adversarial Networks (GANs) as part of their approach to generate adversarial examples which made adversarial examples more natural to human. [11] proposed AdvGAN to generate adversarial examples with generative adversarial networks (GANs), which can learn and approximate the distribution of original instances. Once trained, the feed-forward generator can produce adversarial perturbations efficiently. However, they only generate perturbation by add loss to make target model predict wrong, instead of considering the relationship between different adversarial attack methods.

### 2.2 Defense to Adversarial examples with GAN

So far, there are two main ideas to defend against adversarial attack. A more straightforward approach is to make the model more robust by enhancing training data or adjusting learning strategies, such as adversarial training and defensive distillation. The second is a series of detection mechanisms for detecting and rejecting against the examples. One important way in the second method is to utilize GAN to defense adversarial examples[14], with the advantage to learn the latent distribution of perturbation and reconstruct clean samples better. [17] propose a framework for reconstructing images based on GAN using adversarial examples to generate clean samples similar to the original samples. First, the original sample and the adversarial example training are used to generate the GAN. After the training, the adversarial example and the original sample are first passed through the generator to eliminate the adversarial perturbations, and then the target classifier is classified. The author consider that the misclassification of the adversarial examples is mainly caused by some pixel-level intentional imperceptible perturbation of the input image, so it is desirable to propose an algorithm to eliminate the adversarial perturbations of the input image, thereby achieving the purpose of defending against the attack.

### 2.3 Adversarial and Clean Examples Are Not Twins

There are different viewpoints why adversarial examples exist because of the unexplained nature of DNNs[18]–[20].However, it is widely accepted that the linear properties of deep neural networks in high latitude space are sufficient to generate adversarial attack[8]. There is also a guess that adversarial examples and clean samples are subject to two independent distributions[21], which makes style transfer between the two data possible. To use domain adaptation for style transfer, [16] proposed CycleGAN for learning to translate an image from a source domain X to a target domain Y in the absence of paired examples. The training procedure requires a set of style images in the same style and a set of target images of similar content. The learned mapping function takes one target image as input and transforms it into the style domain.

## 3 Integration of Adversarial Attack and Defense

In this section, we will first introduce the problem definition, and then briefly descript two approaches we utilized to generate adversarial images, at last elaborate the framework, formulation and corresponding network architecture of our proposed Cycle-Consistent Adversarial GAN (CycleAdvGAN).

# CycleAdvGAN: integration of adversarial attack and defence

## 3.1 Problem Definition

Let $A \subseteq R^n$ be the clean feature space, with $n$ the number of features. Suppose that $(a_i, t_i)$ is the $i$th instance within the clean dataset, which is comprised of feature vectors $a_i \in A$, generated according to some unknown distribution $a_i \sim P_A$, and $t_i$ the corresponding true class labels.

Let $B \subseteq R^n$ be the adversarial feature space, with $n$ the number of features. Supposing $(b_i, l_i)$ is the $i$th instance within the adversarial dataset, which is comprised of feature vectors $b_i \in B$, generated according to some unknown distribution $b_i \sim P_B$, and $l_i$ the corresponding predict labels.

The learning model aims to mapping functions between two domains A and B given training samples $\{a_i\}_{i=1}^{N}$ where $a_i \in A$ within the clean dataset and $\{b_i\}_{i=1}^{M}$ where $b_i \in B$ within the adversarial dataset. We denote the data distribution as $a_i \sim P_A$ and $b_i \sim P_B$. Given an instance $a_i$, the goal of the Generator A($G_A$) is to generate adversarial example $b_i$, which is classified as $F(b_i) \neq t_i$ (untargeted attack), where $t$ denotes the true label. And given an adversarial examples $b_i$, the goal of the Generator D($G_D$) is to recover $b_i$ to clean instance $a_i$, which is classified as $F(a_i) = t_i$, where $t$ denotes the true label. $b_i$ should also be close to the original instance $a_i$ in terms of $L_2$ or other distance metric.

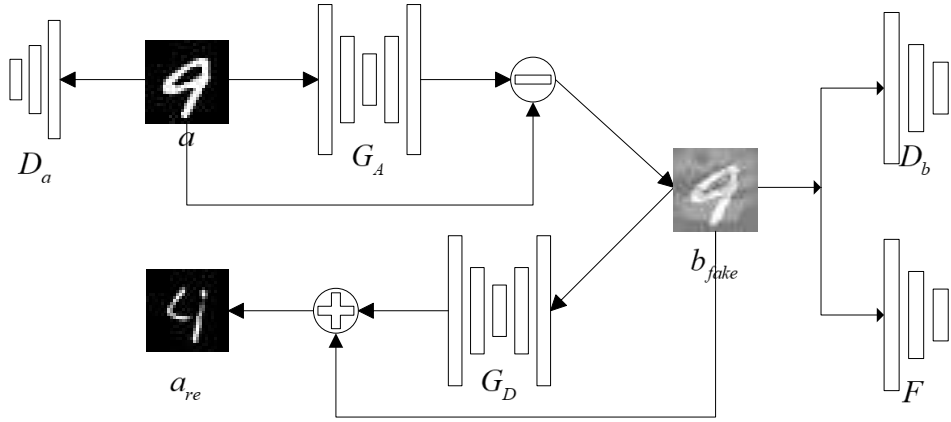

(a) the attack part of the architecture

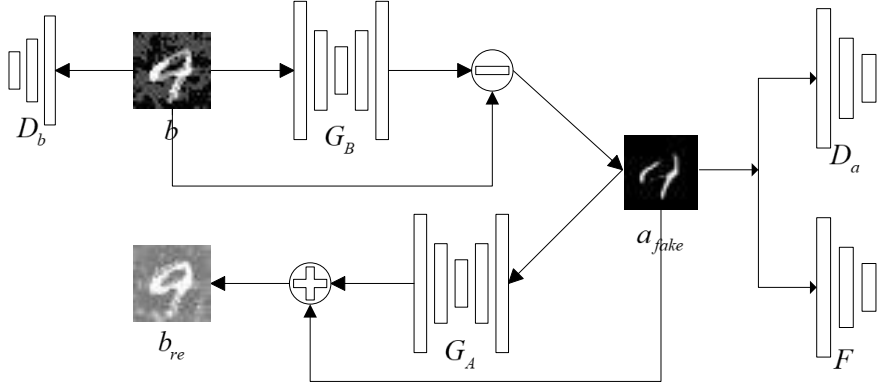

(b) the defense part of the architecture

Figure1. The framework of CycleAdvGAN consists of two generator $G_A$ and $G_D$, two discriminator $D_a$ and $D_b$, and attacking a target model $F$.

## 3.2 Methods Generating Adversarial Examples

In this subsection, two approaches we utilized to generate adversarial examples as the training set are provided with a brief description.

**Fast Gradient Sign Method Attack (FGSM).** Goodfellow et al. proposed a fast method called Fast Gradient Sign Method to generate adversarial examples [8]. They only performed one step gradient update along the direction of the sign of gradient at each pixel. Their perturbation can be expressed as:

$$p = \epsilon sign(\nabla J(\theta, I_c, l))$$

where $\epsilon$ is the magnitude of the perturbation. The generated adversarial example $x_{adv}$ is calculated as: $x_{adv} = x + p$. This perturbation can be computed by using back-propagation. They claimed that the linear part of the high dimensional deep neural network could not resist adversarial examples, although the linear behavior speeded up training. Regularization approaches are used in deep neural networks such as dropout. Pre-training could not improve the robustness of networks.

**Basic Iterative Method (BIM).** Kurakin et al. applied adversarial examples to the physical world [9]. They extended Fast Gradient Sign method by running a finer optimization (smaller change) for multiple iterations. In each iteration, they clipped pixel values to avoid large change on each pixel:

$$I_p^{i+1} = Clip\{I_p^i + \alpha sign(\nabla J(\theta, I_p^i, l))\}$$

where *Clip* limits the change of the generated adversarial image in each iteration.

## 3.3 The CycleAdvGAN Method

Figure 1 illustrates the overall architecture of CycleAdvGAN including two mappings $G_A: A \rightarrow B$ and $G_D: B \rightarrow A$, which is mainly consists of five parts: the generator $G_A$, the generator $G_D$, the discriminator $D_a$, the discriminator $D_b$, and the target neural network $F$. Here the generator $G_A$ takes the original instance $a$ as its input and generates perturbation $G_A(a)$. Then $b_{fake} = a + G_a(a)$ will be sent to the discriminator $D_b$, which is used to distinguish the generated data and the adversarial example $b$. $D_b$ encourages $G_A$ to translate $A$ into outputs indistinguishable from domain $B$. As shown in Figure 1(a) and (b), the two part of the architecture are symmetrical. Therefore, the generator $G_D$ takes the adversarial example $b$ as its input and generates a perturbation $G_D(b)$. Then $a_{fake} = b + G_D(b)$ will be sent to the discriminator $D_a$, which is used to distinguish the generated data and the original instance $a$. $D_a$ encourages $G_B$ to translate $B$ into outputs indistinguishable from domain $A$. To fulfill the goal of fooling the learning model, we perform the white-box attack, where the target model is $F$ in this case. $F$ takes $b_{fake}$ and $a_{fake}$ as its input and outputs its loss $L_{adv}$, which represents the distance between the prediction and the target class $t$ (targeted attack), or the opposite of the distance between the prediction and the ground truth class (untargeted attack).

**Adversarial loss for $G_A$.** We apply adversarial losses to both mapping functions. For the mapping function $G_A : A \to B$ and its discriminator $D_a$, we express the objective as:

$$L_{G_A}(G_A, D_B) = E_{b \sim P_B}[log(D_B(b))] + E_{a \sim P_A}[log(1 - D_B(G_A(a)))]$$

where generator $G_A$ aims to generate imperceptible perturbation $G_A(a)$ that added to $a$ for looking similar to original instance, while $D_b$ aims to distinguish between generated adversarial example and adversarial example.

**Adversarial loss for $G_B$.** We introduce a similar adversarial loss for the mapping function $G_D : B \to A$. The loss is defined as:

$$L_{G_B}(G_D, D_B) = E_{a \sim P_A}[log(D_A(a))] + E_{b \sim P_B}[log(1 - D_A(G_D(b)))]$$

where generator $G_B$ aims to recover adversarial example $b$ to clean example, while $D_a$ aims to distinguish between generated clean example and clean example.

**Cycle Consistency Loss.** After competing with discriminator $D_a$ in the minmax game, $G_A$ should be able to generate visually realistic adversarial examples. However, since no pairwise supervision is provided, the deblurred image may not retain the content information in the original blurred image. Inspired by CycleGAN for style transfer which is shown to imply the cycle-consistency constraints, we also introduce variants of cycle consistency losses to ensure the successful of integration of adversarial attack and defense. The adversarial loss constrains the output of G after added to input to look like the instance in transfer domain. However, adversarial losses alone cannot guarantee that the learned function can map an individual input $a$ to a desired output $b$. To further generate the specific perturbation, we attempt to add cycle consistence loss to the architecture. The loss is defined as:

$$L_{cyc} = E_{a \sim P_A}[\| G_D(G_A(a)) - a \|_1] + E_{b \sim P_B}[\| G_A(G_D(b)) - b \|_1]$$

**Adversarial loss for $F$.** The loss for fooling the target model f in an untargeted attack is:

$$L_{adv} = E_{a \sim P_A}[L_F(b_{fake}, l_t)] + E_{b \sim P_B}[L_F(a_{fake}, l_c)]$$

where $l_t$ is the target label and $l_c$ represents the true class. Meanwhile, $L_F$ denotes the loss function (e.g., cross-entropy loss) used to train the original model $F$. The $L_{adv}$ losses encourages the perturbed image to be misclassified and the recovered instance to be correctly classified.

**Total Loss.** We optimize a min-max objective function $\min_{G_A, G_B, F} \max_{D_A, D_B} L$, where the loss $L$ is defined as:

$$L = \lambda_1 L_{D_A} + \lambda_2 L_{D_B} + \lambda_3 L_{cyc} + \lambda_4 L_{adv}$$

$\lambda_1$, $\lambda_2$, $\lambda_3$ and $\lambda_4$ are the weights to balance the multiple objectives. The next section will provide more training details and discuss the appropriate weights.

## 4  Implementation

**Network Architecture.** Next, we briefly introduce the network architectures in our CycleAdvGAN framework. We adopt the variant architecture for our generator and discriminator from Zhu et al. who have shown impressive results for neural style transfer in the absence of paired examples.

**Generator.** This generator network contains two stride-2 convolutions, several residual blocks, and two strided-2 deconvolutions. we employ the Resnet architecture in our generator, which allows low-level information to shortcut across the network, leading to better results. We use 4 blocks for 28×28 Lenet5 images and 4 blocks for 32×32 cifar10. The encoder-decoder architecture consists of:

Encoder: C8-C16-C32

Decoder: C32-C16-C8

**Discriminator.** For the discriminator networks, we use three stride-2 convolutions. The last layer of discriminator network is fed into a linear layer to generate a 1-dimensional output, followed by a Sigmoid function. Our discriminator architecture is:

C8-C16-C32

| A | B |
|---|---|
| Conv1(32,3,3)+Relu | Conv1(32,3,3)+Relu |
| Max Pooling | Conv1(32,3,3)+Relu |
| Conv1(64,3,3)+Relu | Max Pooling |
| Max Pooling | Dropout(0.5) |
| FC(200)+Relu | FC(128)+Relu |
| FC(10) | Dropout(0.5) |
|  | FC(10) |

Table 1. Target model architectures for the MNIST. A are the LENET5.B are the LENET5 with dropout.

**Target Model.** For the target model, we trained different models on MNIST[22] and CIFAR10[23] respectively. For MNIST, in all of our experiments, we generate adversarial examples for three models whose architectures are shown in Table1. For CIFAR-10, we select ResNet-18[24] and VGG-16[25] for our experiments. We show the classification accuracy of pristine MNIST and CIFAR10 test data in Table 2. The targeted models $F$ could be any given deep networks with the last two layers accessible (e.g., Soft-max layer and the layer before it). These two layers are used as a part of $L_{adv}$. To perform adversarial attack, the loss $L_{adv}$ encourages the adversarial example $b$ to be misclassified by $F$ and the clean examples $a$ to be correctly classified by $F$.

| Model | MNIST | | CIFAR-10 | |
|---|---|---|---|---|
| | A | B | Resnet-18 | VGG-16 |
| Classification Accuracy | 98.85 | 99.14 | 85 | 84 |

Table 2. Classification accuracy rates (%) of benign input for different target models trained by us on MNSIT and CIFAR10.

**Training details.** Our code and models will be available upon publication. We apply the loss in Carlini & Wagner [10]as our loss $L_{adv}^F = \max(\max_{i \neq t} F(x_A)_i - F(x_A)_t, k)$, where $t$ is the target class, and $F$ represents the target network in the semi-white box setting. We set the confidence $k=0$ and use Adam as our solver [26]. For $L_{GAN}$, as same as the Zhu et.al, we replace the negative log likelihood objective by a least-squares loss. This loss is more stable during training and generates higher quality results. In particular, for a GAN loss $L_{GAN}(G, D, A, B)$, we train the G $E_{a \sim P_A}[(D(G(x))-1)^2]$ to minimize and train the D to minimize $E_{b \sim P_B}[(D(b)-1)^2] + E_{a \sim P_A}[D(G(x))^2]$.

**Implementation Details.** In our experiments, we use Pytorch for the implementation and test them on a NVIDIA Tesla V100 GPU cluster in Nvidia DGX station. We train CycleAdvGAN for 100 epochs with a batch size of 64, with the learning rate of 0.01, decreased by 10% every 20 steps.

## 5 Experimental results

In this section, we first evaluate CycleAdvGAN for both semi-white box and black-box settings on MNIST and CIFAR10. We then apply CycleAdvGAN to generate adversarial examples on different target models and test the attack success rate for them. Meanwhile we also recover adversarial examples to clean instances and test the recovery success rate for them and show that our method can achieve higher attack success rates as well as the higher recovery success rates. We use the classification accuracy to measure attacking performance and defending performance with the lower accuracy indicating better attacking performance and the higher accuracy indicating better defending performance. We generate all adversarial examples for different attack methods based on the $L_\infty$ bound as 0.3 on MNIST and 0.03 on CIFAR10.

### 5.1 CycleAdvGAN in semi-white box setting

In semi-white box condition, there is no need to access the original target model after the generator is trained, in contrast to traditional white-box condition. First, we apply different architectures for the target model $F$ for MNIST and with ResNet-18 and VGG-16 for CIFAR10. We apply CycleAdvGAN to perform semi-white box setting against each model on MNIST dataset and CIFAR10 dataset. From the performance of semi-white box setting (classification accuracy rate) in Table 3, we can see that CycleAdvGAN is able to generate adversarial instances to attack all models with high attack success rate compared to other state-of-the-art attacks. Meanwhile, from the performance of semi-white box condition (defense with $G_D$) in Table 3, we can also see that CycleAdvGAN is able to recover adversarial examples to clean samples, so as to significantly improve classification accuracy rates. Efficiently proved by this, our proposed CycleAdvGAN can achieve the integration of adversarial attack and defense.

As shown in the Table 3, we further analyze that in the MNIST dataset, it can be seen that B architecture has better resistance against adversarial attack, especially for FGSM method. This is because dropout [27] was added to B architecture, which can enhance the robustness of neural network. Attacking ability and defending ability is highly correlated with the capacity of the learning model generating the adversarial examples. For example, adversarial generated by B show good attacking performance on A and after defending with $G_D$, A has a better classification accuracy performance, because B usually owns more complicated network structure with dropout and thus better capability than A in practice. And our proposed method can greatly increase the attack rate, because our network learns the potential distribution of adversarial examples. Meanwhile, it can be seen that the CycleAdvGAN trained by the training set of adversarial examples crafted by FGSM and BIM have a higher attack rate than the original attack method through its attack mechanism, and can greatly improve its classification accuracy rate through the defense mechanism.

| Method | MNIST | | CIFAR-10 | |
|---|---|---|---|---|
| | A | B | Resnet-18 | VGG-16 |
| FGSM | 6.12 | 10.53 | 10.27 | 13.2 |
| FGSM (attack with $G_A$) | 2.54 | 1.27 | 5.2 | 3.6 |
| FGSM (defense with $G_D$) | 98.12 | 92.45 | 54.8 | 43.8 |
| BIM | 0.76 | 1.1 | 8.96 | 8.97 |
| BIM (attack with $G_A$) | 0.6 | 0.43 | 1.98 | 2.6 |
| BIM (defense with $G_D$) | 94.6 | 93.15 | 44.8 | 38.6 |

Table 3. Classification accuracy rates of adversarial examples crafted by FGSM, BIM and CycleAdvGAN. FGSM (attack with $G_A$) and FGSM (defense with $G_D$) means that we train the CycleAdvGAN with the adversarial examples crafted by FGSM. The lower accuracy indicates better attacking performance. The higher accuracy indicates better defending performance.

## 5.2 CycleAdvGAN in black-box setting

In this section, we evaluate the transferability of the CycleAdvGAN in the black-box attacking settings. In black-box attacks, we train the CycleAdvGAN in certain target models and optimize the generator accordingly. Once the CycleAdvGAN trained, we generate adversarial examples or recover them through the $G_a$ and $G_D$. Table 4 and Table 5 respectively shows the classification accuracy of MNIST and CIFAR10 datasets, when transferring attacks between different classification models. The transferability performance is marked red in Table 4 in MNIST dataset, from which we can get the following conclusions:

- Adversarial examples generated by CycleAdvGAN have very encouraging transferability among different target models, which means attack by our CycleAdvGAN can perform quite well in black-box setting.
- After defending with $G_D$, the classification accuracy rates have been significantly improved even source and target model are different, which means defense by our CycleAdvGAN also can perform quite well in black-box setting.
- From the above two points, our proposed CycleAdvGAN method can effectively be applied to practical.

| Source / Target | | FGSM | BIM | $G_A$ (FGSM) | $G_A$ (BIM) |
|---|---|---|---|---|---|
| B | Without $G_D$ | 46.530 | 28 | 34 | 24 |
| B | Defense with $G_D$ | 98 | 97 | * | * |

(Source: A)

Table 4. Classification accuracy of adversarial examples generated by different methods transferred between different models on MNIST. $G_A$ (FGSM) means that we train the CycleAdvGAN with the adversarial examples crafted by FGSM.

CycleAdvGAN: integration of adversarial attack and defence

| Target<br>Source | | ResNet-18 | | | |
|---|---|---|---|---|---|
| | | FGSM | BIM | $G_A$ (FGSM) | $G_A$ (BIM) |
| VGG-16 | Without $G_D$ | 32.6 | 19.7 | 23.48 | 13.6 |
| | Defense with $G_D$ | 43.1 | 33.7 | * | * |

Table 5. Classification accuracy of adversarial examples generated by different methods transferred between different models on CIFAR10.

### 5.3 High Transferability of Adversarial Examples Analysis

To interpret why CycleAdvGAN demonstrates better transferability, we further examine the update directions given by BIM and CycleAdvGAN along the iterations. We calculate the cosine similarity of two successive perturbations and show the results in Table 7 when attacking MNIST and CIFAR10 dartasets. The update direction of CycleAdvGAN is more stable than that of BIM due to the larger value of cosine similarity in CycleAdvGAN. Recall that the transferability comes from the fact that models better learn the distribution of adversarial examples rather than perturbation to a certain sample, resulting in better transferability for black-box attacks. Another interpretation is that the GAN can reconstruct images naturally, which may be helpful for the transferability.

| | FGSM | BIM | CycleAdvGAN |
|---|---|---|---|
| cosine similarity | 0.3072 | 0.35 | 0.44 |

Table 6. The cosine similarity of three successive perturbations in FGSM, BIM and CycleAdvGAN. The results are averaged over 10000 images.

### 5.4 The better efficient of Generating adversarial examples

In general, as shown in Table 3, CycleAdvGAN has obvious advantages on the integration of adversarial attack and defense over other white-box and black-box methods. For instance, regarding computation efficiency, CycleAdvGAN performs much faster than others even including the efficient FGSM in attack, although CycleAdvGAN needs extra training time to train the generator. Besides, FGSM and optimization methods can only perform white-box attack, while CycleAdvGAN is able to attack in semi-white box setting.

| | FGSM | BIM | CycleAdvGAN(attack) | CycleAdvGAN(defense) |
|---|---|---|---|---|
| Run time | 1.734s | 18.436s | 0.52s | 0.53s |

Table 7. Comparison with the state-of-the-art attack methods. Run time is measured for generating 1,0000 adversarial instances during test time.

## 6  Conclusion

In this paper, we have proposed CycleAdvGAN to generate adversarial examples and recovery adversarial examples in cycle-consistency. More importantly, we show that the proposed objective is enhancing both the attack effect and the defense effect through the integration of adversarial attack and defense. We show that it can improve attack effect only trained on the adversarial dataset generated by corresponding adversarial attack method.

Further, in our CycleAdvGAN framework, once trained, the generator $G_A$ can produce adversarial perturbations and generator $G_B$ can eliminate adversarial perturbation both efficiently. In addition, the CycleAdvGAN can work with other attacks or defense strategies, not conflict with other existing frameworks. It can also perform both semi-white box and black-box settings with high attack success rate as well as defense rate.

More importantly, we demonstrated that the adversarial and clean data are not twins, subjecting to two different distributions. Consequently, the CycleAdvGAN can better learn the latent distribution of them, instead of targeting special image, so as to improve transferability. Significant transfer performances achieved by our crafted perturbations can pose substantial threat to the deep learned systems in terms of black-box attacking. Therefore, it is an important research direction to be focused on in order to build reliable deep learning systems.